# Leveraging Compliant Tactile Perception for Haptic Blind Surface Reconstruction

1st Laurent Yves Emile Ramos Cheret
*Department of Computer Science*
Lakehead University
Orillia, Canada
lramosc@lakeheadu.ca

2nd Vinicius Prado da Fonseca
*Department of Computer Science*
Memorial University of Newfoundland
St. John's, Canada
vpradodafons@mun.ca

3rd Thiago Eustaquio Alves de Oliveira
*Department of Computer Science*
Lakehead University
Orillia, Canada
talvesd@lakeheadu.ca

*Abstract*—Non-flat surfaces pose difficulties for robots operating in unstructured environments. Reconstructions of uneven surfaces may only be partially possible due to non-compliant end-effectors and limitations on vision systems such as transparency, reflections, and occlusions. This study achieves blind surface reconstruction by harnessing the robotic manipulator's kinematic data and a compliant tactile sensing module, which incorporates inertial, magnetic, and pressure sensors. The module's flexibility enables us to estimate contact positions and surface normals by analyzing its deformation during interactions with unknown objects. While previous works collect only positional information, we include the local normals in a geometrical approach to estimate curvatures between adjacent contact points. These parameters then guide a spline-based patch generation, which allows us to recreate larger surfaces without an increase in complexity while reducing the time-consuming step of probing the surface. Experimental validation demonstrates that this approach outperforms an off-the-shelf vision system in estimation accuracy. Moreover, this compliant haptic method works effectively even when the manipulator's approach angle is not aligned with the surface normals, which is ideal for unknown non-flat surfaces.

This work was supported by the Natural Sciences and Engineering Research Council of Canada (NSERC) - Discovery Grant number RGPIN-2020-04309.

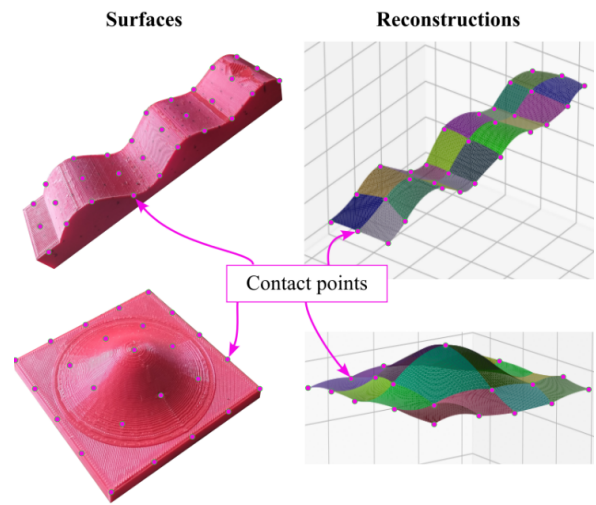

Fig. 1: Blind surface reconstruction using Bézier patches. Contact points shown are evenly spaced by 2cm. On the left, there are two 3D-printed synthetic surfaces. On the right, there are their respective reconstructions.

## I. INTRODUCTION

The performance of robots operating in unstructured environments is dependent upon their ability to accurately comprehend their surroundings, especially when tasked with complex operations, [1], [2]. Accurately estimating an object's geometry is fundamental to dexterous manipulation [3], [4] especially in unstructured settings due to the presence of unknow non-flat objects and surfaces [5]. Given their innate fast image processing, robots operating in structured environments usually use computer vision for three-dimensional surface reconstruction [6]. However, in unstructured settings, such as outdoor or cluttered environments, vision systems encounter more limitations, such as transparent surfaces, reflections, and occlusions [7]. These challenges can lead to errors in reconstructed surfaces, significantly impacting the robot's ability to accurately perform tasks and correctly identify objects [8].

Tactile sensing has long been seen as a powerful source of complementary information to visual systems [9]. Over time, the need has arisen to effectively model surfaces based on sparse tactile data [10], [11]. Graphical models are crucial for reducing the extensive time required for comprehensive tactile exploration while enabling robots to gain an essential understanding of objects and environments. While some approaches rely solely on polynomial fitting of dense contact positions to reconstruct surfaces, others resort to repeated grasping to identify objects from databases [10]–[13]. Nonetheless, the problems of time-consuming probing and complex polynomial fitting are still challenges for reconstructing surfaces in unstructured environments. Recent proposals have leaned towards image-based tactile sensors such as Gelsight [14], [15], which can excel in surface reconstruction of small objects but depend on the alignment of the angle of approach and predetermined surface normals, limiting the range of surfaces to be reconstructed and their applicability in unstructured environments.

Our work introduces an innovative haptic surface reconstruction method that generates a 3D graphical representation while using sparse contact points. Fig. 2 (a) shows how we employ a robotic manipulator's kinematic chain along with a bio-inspired compliant tactile sensing module [16]. The tactile module shown in Fig. 3 comprises magnetic, angular rate, gravity, and barometric sensors. In Fig. 2 (b)-(d), we show how the manipulator moves its end-effector, probing the surface irrespect of the surface local normal and acquires contact information with $2cm$ spacing in the X-Y plane, forming a grid of surface points as seen in Fig. 2 (e) and (f). Figs. 2 (b) and (c) also show how the sensing module deforms to estimate the local normal at the point of contact. The position of control points that define the curvature between contact points is estimated in a geometrical approach and added to the grid of surface points shown in Fig. 2 (g)-(h). Fig. 2 (i) shows how we construct Bézier surface patches using this information. These patches combine to create the overall surface estimation shown in Fig. 2 (j).

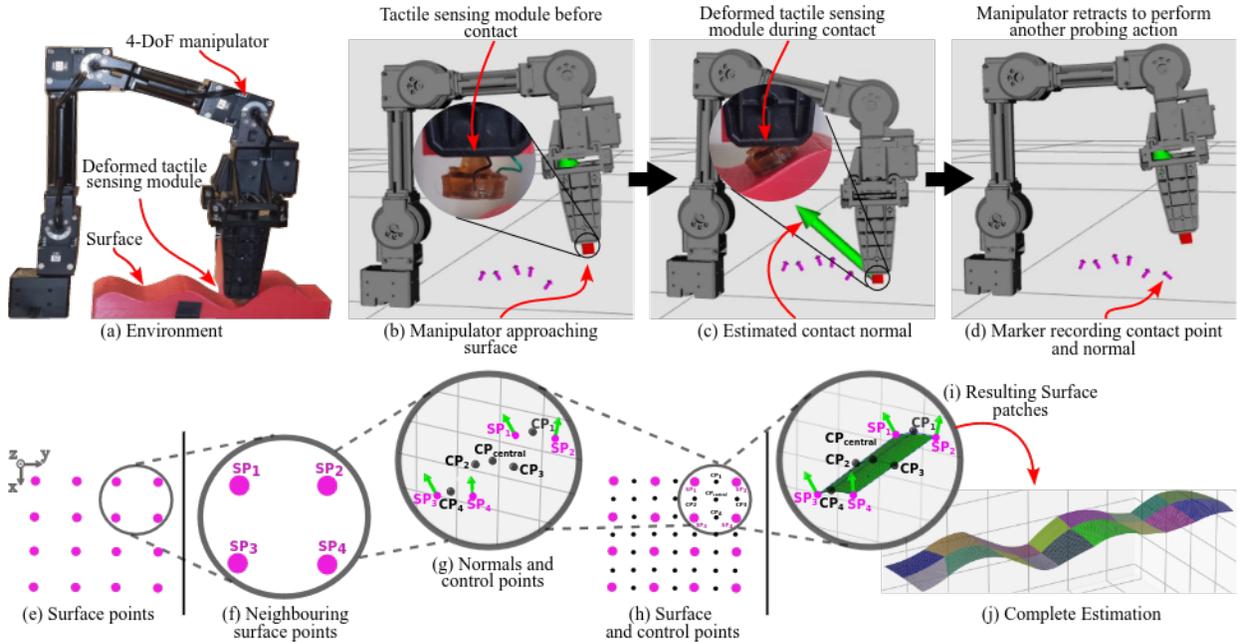

Fig. 2: (a) Shows the Environment composed of 4-DoF manipulator, tactile sensing module, and surface; (b-d) Shows the trajectory performed by the manipulator to probe the surface; (e-h) show how surface points and normals are used to estimate control points; (i,j) shows the resulting NURBS surface patch and a combination of patches to form a complete estimation.

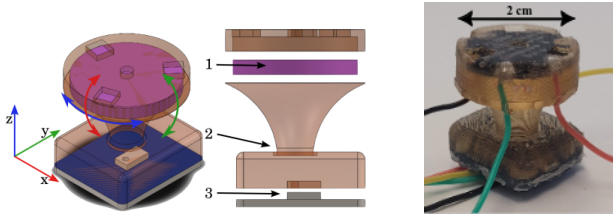

Fig. 3: Tactile sensing module. Components: 1—MARG (magnetic, angular rate, and gravity) system; 2—compliant structure; 3—barometer [16]–[18].

In summary, the present work contributions are:
- We introduce an innovative technique for haptic surface reconstruction, generating 3D graphical models from sparse contact points and normals.
- This approach maintains its effectiveness even if the end-effctor's approach angle deviates from the surface's normal vectors, making it suitable for larger surfaces and unstructured environments.
- We efficiently estimate the local curvature of patches, thereby reducing the time-consuming probing process.
- Larger surfaces can be reconstructed with more patches without increasing mathematical complexity.

To assess the accuracy of our reconstructions, we fabricated five synthetic 3D-printed surfaces with different shapes and sizes, which can be estimated using the present method and compared to their original STL models. Cloud points of the five surfaces, acquired using an off-the-shelf visual system, are also compared to the original models, showing more precision from our reconstructions.

The next section reviews existing literature and identifies research gaps. Section III presents the details of our approach. Section IV presents experimental findings and a comprehensive discussion. Finally, the section V summarizes key contributions and outlines future directions.

## II. RELATED WORKS

Recent work has highlighted the benefits of contact point-based approaches recognizing/reconstructing surfaces, for manipulation and exploration [2], [9].

In the early work by De Oliveira et al. [19] the authors demonstrate the use of MEMs MARGs in tactile sensing. These sensors, combined with kinaesthetic data from motors, enable the effective identification of surface profiles during rigid and soft contact. This approach highlights the potential of multimodal sources in enhancing tactile perception capabilities.

In [20], the authors present the use of a multimodal tactile sensing module to estimate the inclination of a surface around contact points and how these points can be used to perform surface approximation tasks using linear and cubic interpolations as well as Bézier curves. Their approach presented the advantages of directly estimating the contact normal.

Past methods often required dense point sets, leading to time-consuming probing [21]–[23]. This trend is also seen in object recognition, where repeated grasping has shown high accuracy in land [12] and underwater environments [13], but requires a large number of repetitions. Ilonen et al. [24] combine vision and grasping for reconstructions. However, they face the repeated grasping challenge by imposing a symmetry constraint. Estimating surface patches can reduce this time-consuming step, making it more suitable for unstructured environments.

Initial approaches relied on approximating contact point clouds to primitive superquadric functions [10], [11], but these approaches face issues related to mathematical complexity. Casselli et al. [25] instead used primitive polyhedral shapes. Liu and Hasegawa [26] used a strategy that fits triangular B-spline patches to a set of contact points and normals, forming whole surfaces. However, their approach only averages normal vectors for the direction of the resulting flat patch, requiring a dense set of contact points.

Recent work on tactile surface reconstruction largely focused on image-based tactile sensors for high precision. Lu et al. [27] combined high-precision tactile images from different probing locations to face the limitations in the scalability of GelSight modules [14] but were limited by the constant angle of approach and knowledge of the textures. The present work does not rely on visual information, large Deep Learning architectures, or previous knowledge of textures and can approach a surface from different angles.

## III. METHODOLOGY

The setup consists of a 4-DoF manipulator with a tactile sensing module interacting with a surface (see Fig. 4). Real-time orientation and position are determined using MARG sensor data and manipulator inverse kinematics.

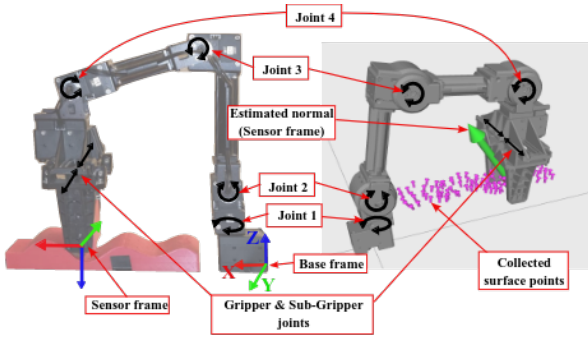

Fig. 4: Views of the robotic manipulator joints. Joints 1 to 4 rotate to provide 4-DoF. Gripper and Sub-Gripper joints hold the sensing module.

The flexible frame in the sensing module enables estimating surface orientation and position at each point of contact without requiring a constant approach angle. Using this information, we calculate control points that define local curvature for Non-Uniform Rational B-Splines (NURBS) patches, which collectively form the surface.

### A. Orientation Estimation

Given a set of $K$ contact points $\{p_i \, | \, p_i \in \mathbb{R}^3\}_{i=1}^K$, the respective orientation set of quaternions is given by $\{q_i \, | \, q_i \in \mathbb{H}\}_{i=1}^K$, where $\mathbb{H} = \{w + xi + yj + zk \, | \, w, x, y, z \in \mathbb{R}\}$.

To determine the orientation of the sensing module, an approximation is created using the accelerometer readings. Let $\vec{a_s} = [a_x, a_y, a_z]$ be the normalized accelerometer information in the sensor frame of reference. In the earth frame of reference (base frame), the unitary vector representing the vertical force can be approximated to $\vec{g} = [0, 0, 1]$. Fig. 4 shows the robot joints and relevant frames of reference. The quaternion representing double the intended rotation from $\vec{g}$ to $\vec{a_s}$ is:

$$q = [q_w, q_x, q_y, q_z]$$
$$q_w = \vec{g} \cdot \vec{a_s} \quad [q_x, q_y, q_z] = \vec{a_s} \times \vec{g} \quad (1)$$

The intended rotation quaternion $q_g^{a_s}$ is achieved by interpolation with the zero-rotation quaternion $q_0 = [1, 0, 0, 0]$ after normalization [28]:

$$\hat{q}_g^{a_s} = q + q_0 \quad q_g^{a_s} = \frac{\hat{q}_g^{a_s}}{\|\hat{q}_g^{a_s}\|} \quad (2)$$

Initially, this approach doesn't consider the robot's Z-axis rotation of the base link. However, this rotation is considered later and is based on the base link frame seen in Fig. 4. We determine the rotation angle $\theta_z$ around the Z-axis of the base frame using the robot's inverse kinematic chain between the base frame and the sensor frame, governed by Joint 1. A quaternion representing the rotation of $\theta_z$ degrees around the Z-axis is created:

$$q_{rot} = \left[\cos\left(\frac{\theta_z}{2}\right), 0, 0, \sin\left(\frac{\theta_z}{2}\right)\right] \quad (3)$$

The final orientation is a composition of both rotation quaternions $q_{rot}$ and $q_g^{a_s}$, as their Hamiltonian product:

$$q_i = q_{rot} \otimes q_g^{a_s}. \quad (4)$$

Once the first estimate is generated, it is adjusted using a Madgwick complementary filter that takes into consideration an initial orientation estimation $q_i$, accelerometer $\vec{a}_s$, and gyroscope $\vec{\omega}_s$ to minimize the objective function $f$:

$$f(q_i, g, a_s) = q_i^* \otimes g \otimes q_i - a_s$$
$$f_g(q_i, a_s) = \begin{bmatrix} 2(q_x q_z - q_w q_y) - a_x \\ 2(q_w q_x + q_y q_z) - a_y \\ 2\left(\frac{1}{2} - q_x^2 - q_y^2\right) - a_z \end{bmatrix} \quad (5)$$

The quaternion is updated as follows:

$$q_{i,t} = q_{i,t-1} + \dot{q}_{i,t}\Delta t$$
$$q_{i,t} = q_{i,t-1} + \left(\dot{q}_{\omega,t} - \frac{\nabla f}{\|\nabla f\|}\right)\Delta t \quad (6)$$

$$\nabla f = J_g^T(q_{t-1}) f_g(q_{t-1}, a_s)$$
$$\dot{q}_{\omega,t} = \frac{1}{2} q_{t-1} \otimes w_s \quad (7)$$

Where $\Delta t$ is the time interval between consecutive sensor readings. The Jacobian is given by:

$$J_g(q_i) = \begin{bmatrix} -2q_y & 2q_z & -2q_w & 2q_x \\ 2q_x & 2q_w & 2q_z & 2q_y \\ 0 & -4q_x & -4q_y & 0 \end{bmatrix} \quad (8)$$

### B. Calibration

Vibrations from the robotic manipulator and external factors make it necessary to calibrate the sensor. The accelerometer and gyroscope readings are averaged as $\vec{a}_{s,0}$ and $\vec{\omega}_{s,0}$. We then calculated the error for both accelerometer and gyroscope as $\vec{\epsilon}_{acc}$ and $\vec{\epsilon}_{gyr}$ to subtract from all subsequent readings. Pure quaternions $a$, $\omega$ are extended from vectors $\vec{a}$ and $\vec{\omega}$ in the hamiltonian products and $q^*$ represents the conjugate of $q$.

$$\vec{a}_{b,0} = q_s^{b*} \otimes a_{s,0} \otimes q_s^b$$
$$\vec{\epsilon}_{acc,0} = \vec{a}_{b,0} - \vec{g} \quad (9)$$

$$\vec{\omega}_{b,0} = q_s^{b*} \otimes \omega_{s,0} \otimes q_s^b$$
$$\vec{\epsilon}_{gyr,0} = \vec{\omega}_{b,0} \quad (10)$$

Once the errors are averaged in the base frame, they can be rotated at any instant $t$ using the estimation $q_t$ to retrieve the errors in the sensing module frame of reference.

$$\vec{\epsilon}_{gyr,t} = q_t^* \otimes \epsilon_{gyr,0} \otimes q_t$$
$$\vec{\epsilon}_{acc,t} = q_t^* \otimes \epsilon_{acc,0} \otimes q_t \quad (11)$$

The adjusted accelerometer and gyroscope readings are then given by:

$$\vec{a}_{s,t+1} = \vec{a}_{s,t} - \vec{\epsilon}_{acc,t} \quad (12)$$
$$\vec{\omega}_{s,t+1} = \vec{\omega}_{s,t} - \vec{\epsilon}_{gyr,t} \quad (13)$$

### C. Control points calculation

Given two surface points $p_1$ and $p_2$ in $\mathbb{R}^3$, and the respective normal vectors $\vec{N}_1$ and $\vec{N}_2$, the 3D planes $plane1$ and $plane2$ can be defined. The objective is to find the control point $cp$ between the points $p_1$ and $p_2$. See Figure 5.

$$p'_1 = \text{proj}_{plane2}(p_1) \quad p'_2 = \text{proj}_{plane1}(p_2) \quad (14)$$

$$\begin{aligned} \vec{l}_1 &= p'_2 - p_1 & L_1(t) &= p_1 + t \cdot \vec{l}_1 \\ \vec{l}_2 &= p'_1 - p_2 & L_2(t) &= p_2 + t \cdot \vec{l}_2 \end{aligned} \quad t \in \mathbb{R} \quad (15)$$

The location of the points $m$ and $n$ resulting in the smallest distance between these skew lines $L_1$ and $L_2$, is found: The projection of the middle point on the vertical plane $V$ passing through $p_1$ and $p_2$ gives us the location of $cp$:

$$\begin{cases} m = L_1(t_1) \\ n = L_2(t_2) \\ (m-n)\cdot \vec{l}_1 = 0 \\ (m-n)\cdot \vec{l}_2 = 0 \end{cases} \quad t_1, t_2 \in \mathbb{R}, \quad cp = \text{proj}_{planeV}\left(\frac{m+n}{2}\right) \quad (16)$$

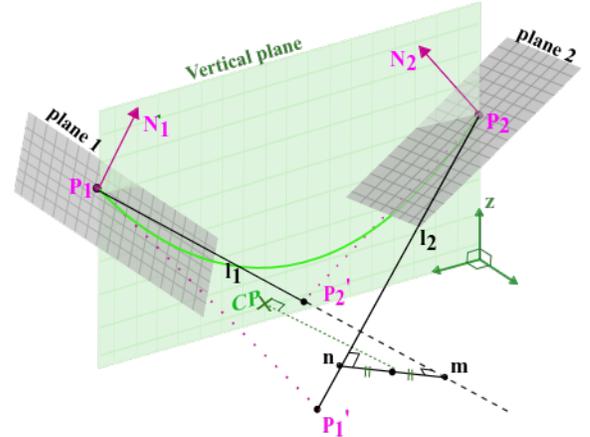

Fig. 5: Control point ($CP$) calculation between surface points $p_1$ and $p_2$. The green line is the generated curvature originating at $p_1$ and finishing at $p_2$.

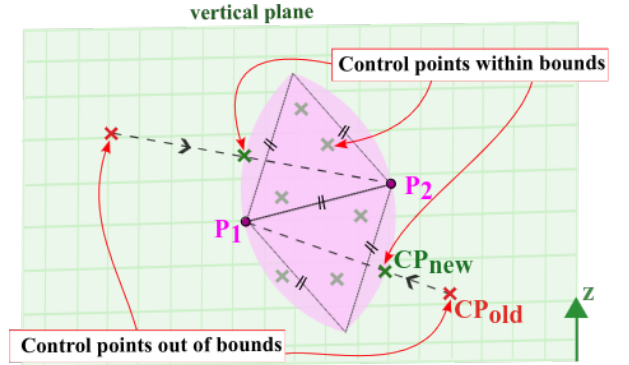

Fig. 6: Adjustment of control points out of bounds ($CP_{old}$) to ($CP_{new}$).

We impose a constraint $|cp-p_1| \wedge |cp-p_2| \leq |p_1-p_2|$ to the distance of any control point $cp$ to its respective surface points $p_1$ and $p_2$. As shown on Fig. 6, the position of the control point out of bounds ($CP_{old}$) is adusted following Algorithm 1 to the new position ($CP_{new}$).

### D. NURBS surface generation

Each surface is formed as a colection of $k$ patches of density $d$: $\mathbb{S} = \{S_i | S_i \in \mathbb{R}^{3\times d}, k \in \mathbb{N}, d \in \mathbb{N}\}_{i=1}^{i=k}$. Each patch $S_i$ is generated from four surface points ($SP_1, SP_2, SP_3, SP_4$) and five control ($CP_1, CP_2, CP_3, CP_4, CP_{central}$), referred to as $P_{ij}$. $CP_{1-4}$ are estimated using $SP_{1-4}$, $CP_{central}$ is an average of the other four. We use quadratic curves (order $k =$

**Algorithm 1** Control point adjustment

**Require:** $p_1, p_2, cp$
  **if** dist($cp, p_1$) $\geq$ dist($cp, p_2$) **then**   ▷ Farthest point to $cp$
    $\vec{l}_{aux} = p_1 - cp$
  **else**
    $\vec{l}_{aux} = p_2 - cp$
  **end if**
  $d \leftarrow |\vec{l}_{aux}|$
  $cp_{new} \leftarrow cp$
  $\delta \leftarrow 0.0001$   ▷ Arbitrary value
  **if** d $\geq |p_1 - p_2|$ **then**
    **while** $(|cp_{new} - p_1| \mathbf{or} |cp_{new} - p_2|) > |p_1 - p_2|$ **do**
      $cp_{new} \leftarrow cp_{new} + \delta \cdot \vec{l}_{aux}$   ▷ Move $CP_{old}$ along $\vec{l}_{aux}$ towards $CP_{new}$
    **end while**
  **end if**

3) in both directions $u$ and $v$.

$$S(u,v) = \frac{\sum_{i=0}^{2}\sum_{j=0}^{2} N_{i,2}(u) \cdot N_{j,2}(v) \cdot P_{ij}}{\sum_{i=0}^{2}\sum_{j=0}^{2} N_{i,2}(u) \cdot N_{i,j}(v)} \quad u,v \in [0,1] \tag{17}$$

We use uniform knot vectors $U = V = [0,0,0,1,1,1]$. The basis function $N(u)$ (and also $N(v)$) of degree p can be defined as:

$$N_{i,0}(u) = \begin{cases} 1, & \text{if } u_i \leq u < u_{i+1} \\ 0, & \text{otherwise} \end{cases} \quad u_i \in U \tag{18}$$

$$N_{i,p}(u) = \frac{u - u_i}{u_{i+p} - u_i} \cdot N_{i,p-1}(u) + \frac{u_{i+p+1} - u}{u_{i+p+1} - u_{i+1}} \cdot N_{i+1,p-1}(u) \tag{19}$$

### E. Experimental setup

The experimental setup was developed using the Robot Operating System (ROS) [29] environment on Ubuntu 16.04. The robotic manipulator used in this work is the OpenMANIPULATOR-X RM-X52-TNM, and robot commands are sent through the embedded board OpenCR 1.0.

The bio-inspired multi-modal sensing module proposed in [16] is used in this work. It comprises an LSM9DS0 MARG (Magnetic, Angular Rate, Gravity) sensor and an MPL115A2 Barometer pressure sensor. These sensors are embedded in a flexible polyurethane, and their structure is shown in Fig. 3. Sensors readings are sent to the main ROS environment through ROS serial publisher at $250k$ baud in a Teensy 3.2 microcontroller. We test our approach by defining five artificial surfaces with different non-flat surfaces. Fig. 7 shows the five surfaces 3D-printed in PLA plastic.

To collect the data on top of the synthetic surfaces, the robotic manipulator vertically moves its end-effector, probing the surface at $2cm$ intervals in a pre-defined motion as previously shown in Fig. 1. Contact between the manipulator and surfaces is registered under loose constraints using the barometer level and joint efforts, and the surfaces are placed to fit the manipulator's pre-defined motion.

### F. Validation

We used the CloudCompare software to compare the 3D CAD files to the robot's estimations. Estimations and ground truth STL are aligned using the Iterative Closest Point (ICP) algorithm. The validation process used two main metrics. We computed the Cloud-to-Cloud (CC) distance between the original surfaces' point cloud and the estimated surfaces' point cloud, and the Cloud-to-mesh (CM) between the original point cloud and the estimated mesh. We use the Hausdorff distance $H(A,B)$ as error metric for the maximum distance from set $A$ to the nearest point in set $B$ in equation:

$$\begin{aligned} H(A,B) &= \max\{h(A,B), h(B,A)\} \\ h(A,B) &= \max_{a \in A} \min_{b \in B} |a-b| \end{aligned} \tag{20}$$

**Cloud-to-Cloud (CC):** For every partition (octree structure), a subset $A_i$ of the reconstruction $A$ is aligned with a subset $B_i$ of the original cloud $B$ to generate $(CC)_i$. For every octree structure $i$ there are two points $a_i$ in the reference cloud and $b_i$ in the mesh that satisfy $H(A_i, B_i)$.

$$CC = \sum_{i=1}^{k} \frac{|\vec{CC_i}|}{k} \quad \vec{CC_i} = H(A_i, B_i) \cdot \vec{u}_i$$
$$\vec{u}_i = \frac{b_i - a_i}{|b_i - a_i|} \tag{21}$$

**Signed (sCM) and Unsigned Cloud-to-Mesh (uCM):** Using the same idea as the previous metric. Let $\vec{n}_i$ be the unitary normal vector to the mesh at $b_i$:

$$\begin{aligned} sCM &= \left|\sum_{i=1}^{k} \frac{s\vec{CM_i}}{k}\right| \quad s\vec{CM_i} = H(A_i, B_i) \cdot \vec{n}_i \\ uCM &= \sum_{i=1}^{k} \frac{|s\vec{CM_i}|}{k} \end{aligned} \tag{22}$$

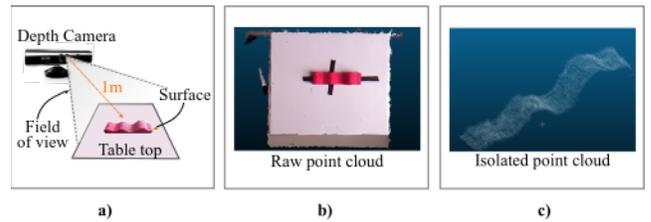

Fig. 8: a) Kinect setup. b) Raw point cloud from field of view. c) Isolated surface point cloud.

We compared our approach to a visual system using a Kinect sensor, equipped with a 640×480 pixels 30 Hz RGB camera and a 640×480 pixels 30 Hz Infra-Red depth camera. Fig. 8 shows the experimental setup used to collect data for the five surfaces from the visual system. We also used CloudCompare to isolate the surface point cloud from the background and calculated metrics, mirroring the procedure for tactile estimations.

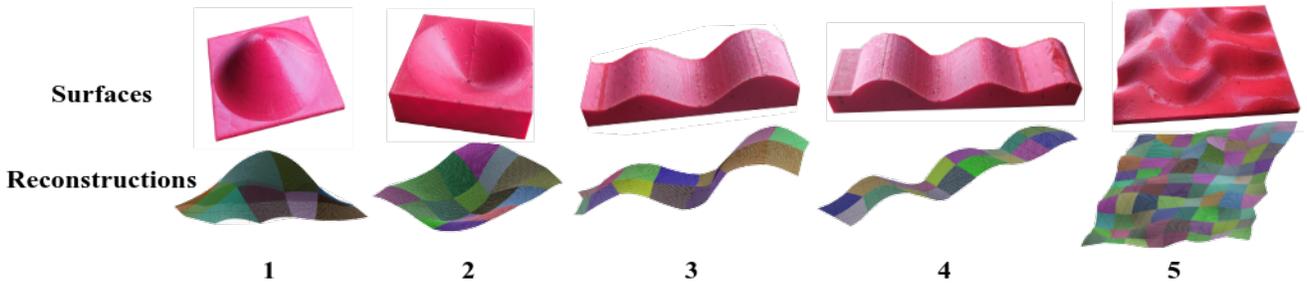

Fig. 7: The five synthetic surfaces under study. Surface 1 is $8cm \times 8cm$ with $3cm$ peak height. Surface 2 is $8cm \times 8cm \times 2cm$ with the lowest height at $1cm$. Surface 3 is $16cm \times 5cm$ with the highest point at $2.5cm$ and the lowest at $1cm$. Surface 4 is $19cm \times 4cm$ with the highest point at $2.5cm$ and the lowest at $1cm$. Surface 5 is $20cm \times 16cm \times 1cm$.

## IV. Results & Discussion

Results from our approach using tactile data are shown in Table I, and those from the vision approach using the Kinect are presented in Table II. The comparison between estimations and original shapes is shown in Fig. 7.

Our method's average $uCM$ distance is 0.652 mm, and $CC$ is 0.637 mm. The $sCM$ has significantly lower values than the $CC$ and $uCM$. This shows that there is no apparent bias in the final estimation.

| Surfaces | $uCM$ | $sCM$ | $CC$ |
|---|---|---|---|
| 1 | $0.75 \pm 0.53$ | $0.017 \pm 0.93$ | $0.75 \pm 0.54$ |
| 2 | $0.70 \pm 0.52$ | $0.011 \pm 0.90$ | $0.69 \pm 0.48$ |
| 3 | $0.53 \pm 0.42$ | $0.014 \pm 0.95$ | $0.49 \pm 0.3$ |
| 4 | $0.72 \pm 0.57$ | $0.012 \pm 0.88$ | $0.67 \pm 0.60$ |
| 5 | $0.55 \pm 0.42$ | $0.003 \pm 0.75$ | $0.59 \pm 0.44$ |

TABLE I: Distances between original surface STL and tactile reconstructions. Average distances and standard deviation in millimeters (mm).

| Surfaces | $uCM$ | $sCM$ | $CC$ |
|---|---|---|---|
| 1 | $1.05 \pm 0.92$ | $0.12 \pm 1.13$ | $1.02 \pm 0.94$ |
| 2 | $0.98 \pm 0.89$ | $0.12 \pm 1.02$ | $0.99 \pm 0.88$ |
| 3 | $0.96 \pm 0.82$ | $0.15 \pm 1.31$ | $1.03 \pm 0.91$ |
| 4 | $1.02 \pm 0.8$ | $0.17 \pm 1.25$ | $1.07 \pm 0.94$ |
| 5 | $1.23 \pm 0.97$ | $0.21 \pm 1.127$ | $1.26 \pm 1.07$ |

TABLE II: Distances between original surface STL and visual reconstructions. Average distances and standard deviation in millimeters (mm).

We compare our results to the Kinect approach, shown on Fig. 9, and notice an increase in the Unsigned $uCM$ and $CC$ distances. The average increase is 60% in comparison to the tactile approach. $sCM$ also shows a significant increase of more than 10 times.

In [23], the authors reported results that show distances from the generated meshes to the surface ranging from 0.1 mm to 0.23 mm to the three surfaces under study (shell, pear, and dark pebble). Nonetheless, they use a greater number of tactile data points (between 97 and 63 points) to cover significantly smaller surfaces (around $2 \times 2$ cm) than those tested here. In comparison to [26], our work also uses a significant smaller number of contact points and normals. In summary, our work uses much less probing (on average 1 point per 2cm) when compared to while also generating a graphic model or mesh.

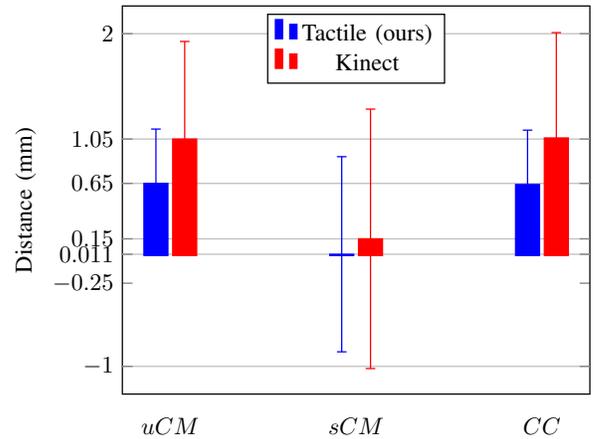

Fig. 9: Distance comparison between our approach and vision using Kinect. Mean and standard deviation.

## V. Conclusion

In conclusion, this paper presents a novel approach for blind surface reconstruction using a robotic manipulator and a compliant tactile sensing module. This method has the advantage of probing surfaces even when the manipulator's angle of approach isn't aligned with surface normals, a crucial advantage over non-compliant tactile systems. Our method reduces the number of probing attempts, a time-consuming step in tactile surface reconstruction, by leveraging the manipulator's kinematic data and the flexible sensing module deformation to estimate control points for spline-based surface patches. Experimental reconstructions on five synthetic surfaces show its effectiveness, with lower estimation errors than a vision system. It also has potential applications in computer vision, overcoming challenges like surface transparency, reflections, and occlusions. The present study uses a single sensing module for probing but can be scaled to use multiple ones in parallel as they are compact and cost-efficient. Future work will explore dynamic tactile exploration and integration with visuotactile systems for dexterous manipulation, promising advancements in this field.